\documentclass[11pt]{article}
\usepackage{float}

\usepackage[final]{acl}

\usepackage{times}
\usepackage{latexsym}

\usepackage[T1]{fontenc}

\usepackage[utf8]{inputenc}
\usepackage{amssymb}
\usepackage{microtype}

\usepackage{inconsolata}

\usepackage{graphicx}
\usepackage{amsmath}
\usepackage{enumitem}
\usepackage{multirow}
\usepackage{booktabs}
\usepackage{colortbl}
\usepackage[dvipsnames]{xcolor}
\usepackage{tikz}
\usepackage{newunicodechar}
\newunicodechar{✗}{\ding{55}}
\usepackage{pifont}
\usepackage[utf8]{inputenc}
\newcommand{\resTag}[2]{%
  \tikz[baseline=(X.base)] 
    \node[draw=gray,
          rounded corners=2pt,
          inner xsep=4pt,
          inner ysep=2pt,
          text=#1,
          font=\scriptsize\sffamily] (X) {#2};%
}
\newcommand{\HIGH}{\resTag{ForestGreen}{HIGH}}
\newcommand{\MID}{\resTag{RoyalBlue}{MID}}
\newcommand{\LOW}{\resTag{Orange}{LOW}}
\newcommand{\UNDER}{\resTag{BrickRed}{UNDER}}

\definecolor{lightgray}{gray}{0.9}
\definecolor{latinblue}{RGB}{31, 119, 180}
\definecolor{cyrilred}{RGB}{214, 39, 40}

\newcommand{\latin}{\textcolor{latinblue}{\large$\bullet$}}
\newcommand{\cyril}{\textcolor{cyrilred}{\large$\bullet$}}

\DeclareUnicodeCharacter{2581}{\texttt{\_}}

\title{Tokenization and Morphological Fidelity in Uralic NLP: \\A Cross-Lingual Evaluation}

\author{Nuo Xu \\
  University of Eastern Finland \\
  \texttt{xnuo@student.uef.fi} \\\And
  Ahrii Kim\thanks{Corresponding author.} \\
  AI-Bio Convergence Research Institute\\
  Soongsil University \\
  \texttt{ahriikim@ssu.ac.kr} \\}

\begin{document}
\maketitle
\begin{abstract}
Subword tokenization critically affects Natural Language Processing (NLP) performance, yet its behavior in morphologically rich and low-resource language families remains under-explored. This study systematically compares three subword paradigms—Byte Pair Encoding (BPE), Overlap BPE (OBPE), and Unigram Language Model—across six Uralic languages with varying resource availability and typological diversity.

Using part-of-speech (POS) tagging as a controlled downstream task, we show that OBPE consistently achieves stronger morphological alignment and higher tagging accuracy than conventional methods, particularly within the Latin-script group. These gains arise from reduced fragmentation in open-class categories and a better balance across the frequency spectrum. Transfer efficacy further depends on the downstream tagging architecture, interacting with both training volume and genealogical proximity.

Taken together, these findings highlight that morphology-sensitive tokenization is not merely a preprocessing choice but a decisive factor in enabling effective cross-lingual transfer for agglutinative, low-resource languages.

\end{abstract}

\section{Introduction}

Subword tokenization plays a key role in modern multilingual Natural Language Processing (NLP) as it determines how linguistic information is segmented and represented for downstream models. In multilingual settings, tokenizers are typically trained to balance vocabulary efficiency and coverage across many languages, often under strong data imbalance \citep{selvamurugan2025from,downey-etal-2024-targeted,limisiewicz-etal-2023-tokenization,petrov_et_al2023}.

This design choice poses particular challenges for morphologically rich languages, where grammatical functions are encoded through extensive inflection and agglutination. In such languages, inappropriate segmentation can obscure morpheme boundaries, inflate sequence length, and hinder cross-lingual transfer \citep{teklehaymanot-etal-2025-movoc,garcia-etal-2025-exploring,asgari2025morphbpe}, affecting downstream tasks such as part-of-speech (POS) tagging \citep{libovicky-helcl-2024-lexically} and machine translation \citep{kim-kim-2022-vacillating}.

These issues are particularly pronounced in the Uralic language family, which combines highly productive morphology with severe data imbalance across languages spanning a wide spectrum of resource availability—from relatively high-resource languages such as Finnish and Hungarian to severely low-resource languages such as Northern Sami and Komi-Zyrian \citep{chelombitko-komissarov-2024-specialized,tereshchenko2025evaluatingopenaigptmodels}. Recent studies show that standard multilingual tokenizers allocate disproportionately little effective vocabulary capacity to Uralic languages \citep{chelombitko-komissarov-2024-specialized}. While alternative strategies such as Overlap BPE (OBPE) have been proposed that encourage shared subword units between high-resource and low-resource languages \citep{patil-etal-2022-overlap}, existing work typically considers tokenization strategies in isolation or on limited language pairs, leaving it unclear how different approaches compare across the morphological and resource diversity of the Uralic language family.

In this work, we conduct a systematic comparison of recent and widely adopted tokenization strategies across six Uralic language pairs. We train tokenization models in the Universal Dependencies dataset and evaluate the impact of different tokenization choices on downstream task performance. Through this comparative analysis, we aim to clarify how tokenization strategies interact with morphological complexity and resource imbalance in Uralic languages. We release our codes.\footnote{\url{https://github.com/farfromshallow/Uralic-language-NLP.git}}

Our contributions are threefold:
\begin{itemize}[noitemsep]
    \item We conduct a controlled comparison of three tokenization paradigms—BPE, Unigram, and OBPE—across six Uralic languages spanning high-, mid-, and low-resource conditions, revealing how resource imbalance affects segmentation behavior. 
    \item We present the first systematic evaluation of OBPE for Uralic languages, demonstrating its consistent gains in morphological alignment and cross-lingual transfer efficiency over conventional multilingual tokenizers. 
    \item We link intrinsic segmentation quality with extrinsic POS tagging performance, showing that morphology-aware tokenization leads to measurable improvements in downstream accuracy, particularly under low-resource and typologically related conditions.

\end{itemize}

\section{Related Work}
This section addresses challenges arising from low-resource and morphologically rich languages in multilingual contexts (\S~\ref{subsec:related_multilingual}), and examines how these issues are amplified within the Uralic case (\S~\ref{subsec:related_uralic}).

\subsection{Challenges in Multilingual Tokenization}
\label{subsec:related_multilingual}

Modern multilingual NLP systems rely on data-driven segmentation algorithms such as BPE \citep{sennrich-etal-2016-neural} and the Unigram language model \citep{kudo-2018-subword}. Although BPE can be formally understood as a compression algorithm solving a combinatorial optimization problem \citep{zouhar-etal-2023-formal}, its linguistic consequences are far from neutral. In morphologically rich languages, statistical subword merges often fail to align with true morpheme boundaries \citep{bostrom-durrett-2020-byte, arnett2025}, fragmenting semantic units into opaque sequences \citep{hofmann-etal-2021-superbizarre} and weakening compositional generalization on rare forms \citep{wolleb-etal-2023-assessing}. Studies on Turkish and Finnish show that, for certain architectures, maintaining full word boundaries can outperform subword segmentation in low-resource conditions because excessive fragmentation disperses semantic information \citep{Hu_2025}.

Beyond morphology, tokenization also encodes structural inequities across languages. Standard vocabulary generation procedures maximize global compression efficiency, which implicitly favors high-resource languages that dominate the training corpus \citep{foroutan2025parityawarebytepairencodingimproving}. Consequently, text from low-resource languages is tokenized into longer sequences, amplifying computational costs and degrading model accuracy. To address this, “Parity-aware BPE” \citep{foroutan2025parityawarebytepairencodingimproving} modifies the merge selection criterion to improve compression for the worst-off languages, thereby enhancing cross-lingual fairness . Similarly, OBPE \citep{patil-etal-2022-overlap} adjusts the BPE objective to encourage shared tokens across related languages, trading a small loss in global compression for greater cross-lingual consistency.

\subsection{Tokenization in Uralic Languages}
\label{subsec:related_uralic}
Recent studies show that massively multilingual models struggle to adapt to Uralic languages for two main reasons. 
First, the “curse of multilinguality” reduces per-language capacity as more languages are added \citep{downey-etal-2024-targeted}, and vocabulary coverage remains poor for morphologically rich forms \citep{a-pirinen-2024-keeping, uralicnlp_2019}. Second, cross-lingual transfer is constrained by genealogical distance; effective transfer requires structural similarity, largely absent between Uralic and Indo-European languages \citep{bankula2025crosslinguistictransfermultilingualnlp}.

Recent work has addressed these challenges through two main approaches: statistical vocabulary adaptation and linguistically informed segmentation. In the statistical paradigm, specialized monolingual tokenizers improve compression and lexical coverage for Northern Sámi and Estonian compared to multilingual baselines \citep{chelombitko-komissarov-2024-specialized}, and compact, domain-adapted vocabularies often outperform large generic ones \citep{downey-etal-2024-targeted}. 

In the linguistic paradigm, morphology-aware tokenizers explicitly integrate morphological boundaries into segmentation. \citet{asgari2025morphbpe} introduced \textit{MorphBPE}, which optimizes merges using morphological cues, while \citet{garcia-etal-2025-exploring} showed that morphology-aware segmentation enhances language modeling for agglutinative languages. Cross-linguistic analyses further reveal that tokenizers aligned with morphological units yield more consistent downstream performance \citep{arnett2025}.

Overall, while statistical adaptation enhances efficiency, morphology-aware segmentation better preserves linguistic structure—an aspect this study systematically evaluates across Uralic languages.
\section{Subword Tokenization Framework}
This study evaluates three representative tokenization paradigms—Byte Pair Encoding (BPE), Unigram Language Model, and Overlap-Based BPE (OBPE)—within the context of multilingual and low-resource language modeling. All three methods aim to optimize the trade-off between vocabulary compactness and representational adequacy, yet differ fundamentally in their learning objectives, inference dynamics, and treatment of morphological variation.

\paragraph{Byte Pair Encoding.} 
BPE \citep{sennrich-etal-2016-neural} is an agglomerative algorithm adapted from data compression \citep{10.5555/177910.177914}, which iteratively merges the most frequent adjacent symbol pair $(A,B)$ until the target vocabulary size is reached. Formally, it maximizes the compression utility \citep{zouhar-etal-2023-formal}, reducing total encoded corpus length. By design, BPE produces deterministic segmentations, ensuring stability but often merging across morphological boundaries. While its frequency-driven merges are highly efficient for frequent tokens, this greediness leads to \textit{morphological opacity}, where semantically distinct morphemes are concatenated simply due to frequent co-occurrence \citep{10.1145/3578707}. Such behavior disproportionately harms low-resource languages, where co-occurrence statistics are sparse and biased toward dominant language distributions \citep{chelombitko-komissarov-2024-specialized}. Recent adaptations attempt to mitigate these effects through adaptive merge thresholds and entropy-based pre-tokenization \citep{hu2025entropy}, yet standard BPE remains sensitive to corpus imbalance.

\paragraph{Unigram Language Model.} 
The Unigram tokenizer \citep{kudo-2018-subword} follows a probabilistic, subtractive approach designed to maximize the marginal likelihood of the training corpus. Unlike agglomerative algorithms such as BPE, Unigram begins with a large seed vocabulary $\mathcal{V}_0$ (e.g., all substrings) and iteratively prunes tokens based on their contribution to the overall likelihood:
\begin{equation}
    \mathcal{L}(\mathcal{V}) = \sum_{X \in D} \log \sum_{\mathbf{x} \in \text{Seg}(X)} \prod_{x_i \in \mathbf{x}} P(x_i).
\end{equation}

The optimization alternates between estimating subword usage (E-step) and re-estimating probabilities with pruning (M-step). This procedure implicitly favors morphologically coherent subwords, as meaningful morphemes tend to have higher independent probabilities \citep{bostrom-durrett-2020-byte}. Its stochastic segmentation via \textit{subword regularization} further exposes the model to multiple valid tokenizations of the same surface form, improving robustness in morphologically productive or noisy settings \citep{kudo-2018-subword, vemula-etal-2025-rethinking}. Recent studies confirm that Unigram consistently outperforms deterministic schemes such as BPE in morphologically rich and agglutinative languages \citep{vemula-etal-2025-rethinking}.

\paragraph{Overlap-Based BPE.} 
OBPE \citep{patil-etal-2022-overlap} reformulates BPE’s merge criterion to promote lexical parity in multilingual corpora. In standard multilingual BPE, tokens from high-resource languages dominate due to higher frequency counts, fragmenting low-resource languages into longer sequences. OBPE introduces a dual optimization objective that jointly maximizes compression and cross-lingual overlap by aggregating token frequencies using a generalized mean:
\begin{equation}
\begin{aligned}
\text{Overlap}(L_i, L_h, S)
&= \sum_{k \in S}
\left(\frac{f_{ki}^p + f_{kh}^p}{2}\right)^{1/p}, \\
&\quad p \le 1.
\end{aligned}
\end{equation}
When $p \to -\infty$, the criterion prioritizes merges that maximize the minimum shared token frequency across languages—favoring lexical forms that exist in both corpora. This ensures that vocabulary growth benefits both high- and low-resource languages, preventing representational dominance. Recent work shows that such overlap-aware tokenization reduces sequence length disparity and improves cross-lingual generalization in morphologically rich, low-resource families \citep{karthika2025multilingualtokenizationlensindian, limisiewicz-etal-2023-tokenization}.

\begin{table}[t]
  \centering
  \small
  \resizebox{\columnwidth}{!}{
  \begin{tabular}{l c c r r r}
    \toprule
    \textbf{Language} & \textbf{Code} & \textbf{Resource} & \textbf{Train} & \textbf{Dev} & \textbf{Test} \\
   \midrule
        \multicolumn{6}{l}{\latin{} \textbf{Latin script languages}} \\
        \midrule
        Finnish        & \texttt{fin} & \HIGH  & 6{,}475 & 769 & 809 \\
        Estonian       & \texttt{est} & \HIGH  & 5{,}444 & 833 & 913 \\
        Hungarian      & \texttt{hun} & \MID   &   910   & 441 & 449 \\
        North Sámi     & \texttt{sme} & \LOW   & 1{,}873 & 624 & 625 \\
        \midrule
        \multicolumn{6}{l}{\cyril{} \textbf{Cyrillic script languages}} \\
        \midrule
        Russian        & \texttt{rus} & \HIGH  & 2{,}080 & 589 & 813 \\
        Komi-Zyrian    & \texttt{kpv} & \UNDER &   397   & 133 & 133 \\
    \bottomrule
  \end{tabular}
  }
  \caption{
  Dataset statistics for all languages (unit: sentences). Resource availability is indicated by categorical tags; \textcolor{BrickRed}{\textsc{\small under}} denotes severely under-resourced languages. Hungarian is treated as a mid-resource control due to its relatively small dataset size.
  }
  \label{tab:language_info}
\end{table}

\section{Experiment Setup}

\subsection{Dataset}
We use Uralic treebanks from the Universal Dependencies (UD) v2 collection \citep{nivre-etal-2020-universal}, a standardized framework that provides linguistically consistent annotations for multilingual NLP tasks, including part-of-speech and dependency relations. 

The selected datasets vary widely in size, ranging from 6,475 down to a mere 397 sentences for Komi-Zyrian. In terms of textual domain, the majority of the corpora consist of newswire text. However, significant variations exist: the Komi-Zyrian dataset is derived exclusively from fiction, whereas the high-resource anchor languages (Finnish and Estonian) exhibit broader genre coverage, including blogs, wikis, and social media content.

This severe sparsity, particularly for Komi-Zyrian, imposes strict constraints on model complexity. To account for this disparity, we adopted a dynamic data partitioning strategy. For languages with sufficient resources, we applied the conventional 8:1:1 split for training, development, and testing. For extremely low-resource treebanks, we used a 6:2:2 ratio, following \citet{sheyanova:2017}, to ensure stable evaluation and mitigate lexical overfitting. Although this reduces the amount of training data, it guaranties adequate test coverage and statistical reliability across all resource levels. 

\subsection{Language Pairs}
To evaluate cross-lingual adaptation under different tokenization schemes, we selected high-resource \textit{source} and low-resource \textit{target} languages with varying degrees of linguistic similarity. In all experiments, models were first trained on the source language and subsequently finetuned on the target language. Languages were additionally grouped by script—Latin and Cyrillic—to eliminate confounding factors from mixed writing systems \citep{tufa-etal-2024-unknown}. All selected languages are morphologically rich and predominantly agglutinative, where grammatical information is expressed through extensive suffixation. This property makes them particularly sensitive to subword segmentation and therefore well-suited for analyzing tokenizer effects.

For the Latin-script group (\latin{}), Finnish and Estonian serve as source languages. Both are Uralic languages with highly productive case systems and complex verbal morphology. They are paired with typologically distinct targets to explore different transfer scenarios under the same adaptation protocol. North Sámi, a closely related Uralic language, enables evaluation of transfer between structurally similar but resource-imbalanced languages. Hungarian, by contrast, represents a more distant Ugric branch with mixed agglutinative–fusional morphology and a rich inflectional system. This setup allows us to test whether structural similarity or morphological typology better predicts adaptation effectiveness when lexical overlap is limited.

For the Cyrillic-script group (\cyril{}), we use Russian as the source and Komi-Zyrian as the target. Although genealogically unrelated—Russian being Indo-European and Komi-Zyrian Uralic—they share the Cyrillic alphabet and a history of geographic contact. Russian exhibits fusional morphology, while Komi-Zyrian retains agglutinative Uralic structure with extensive case marking. This pairing isolates the effect of shared script from genealogical similarity, revealing whether orthographic commonality alone facilitates adaptation.

\subsection{Preprocessing}

To ensure consistent alignment between subword tokenization and gold-standard annotations, we adopt a three-stage preprocessing pipeline widely used in subword-level sequence labelling.

\paragraph{Gold-Standard Extraction.} 
We extracted the FORM column from the UD CoNLL-U files to preserve the original token boundaries while omitting metadata. This approach maintains morphological integrity for complex tokens, such as hyphenated compounds (e.g., \texttt{100-aastased}) and abbreviations (e.g., \texttt{4:sta}), ensuring strict consistency with the gold segmentation defined in the treebank \citep{chiarcos_et_al}.

\paragraph{Greedy Alignment.}
We then aligned tokenizer outputs to the gold tokens through character-level greedy matching after Unicode normalization (NFKC). Implementation-specific boundary markers, such as the SentencePiece \verb|▁| (U+2581) and OBPE \texttt{</w>} symbols, were removed prior to alignment, following common alignment strategies used in multilingual parsing pipelines \citep{rosa-marecek-2018-cuni, che-etal-2018-towards}.

\paragraph{First-Subword Tagging.}
Finally, gold token-level labels (e.g., POS tags) were projected to subword sequences using a first-subword tagging scheme \citep{devlin-etal-2019-bert, pires-etal-2019-multilingual}, where each token label is assigned to its first subword while subsequent subwords are padded. This choice may lead to information loss, particularly in Uralic languages, which often express grammatical categories through suffixes. Nonetheless, we consider this normalization necessary to obtain comparable metrics across tokenizers with different levels of granularity. We further assume that, despite this limitation, the effect is alleviated by the bidirectional nature of the BiLSTM encoder.

\subsection{Training Setup}

We trained the standard BPE and Unigram LM tokenizers using the SentencePiece toolkit \citep{kudo-richardson-2018-sentencepiece}. Each tokenizer was trained on monolingual data from its corresponding language pair to avoid cross-lingual contamination.

For OBPE, we adopted the configuration proposed by \citet{patil-etal-2022-overlap} to promote lexical equity across languages. We assigned equal weights ($\alpha = 0.5$) to the compression and overlap objectives, and set the generalized mean exponent to $p = -\infty$ to prioritize merges that maximize the minimum shared token frequency between paired languages. To maintain comparable capacity across all conditions, the vocabulary size was fixed at $|\mathcal{V}| = 5{,}000$ subword units for all models. Although this limit is smaller than standard industrial vocabularies used for high-resource agglutinative languages, it was required by the severe data sparsity of our target datasets. In a pilot study, we observed that scaling the vocabulary to 8,000 operations caused rapid lexical overfitting in the lowest-resource Komi-Zyrian. Consequently, we adopted the 5,000 limit as a strategic compromise to prioritize model generalizability across the diverse resource tiers in our evaluation.

\paragraph{Cross-lingual training setup.}
To evaluate how tokenizer design affects cross-lingual adaptation, we trained all models on a high-resource source language and subsequently finetuned them on each low-resource target language. This setup allows us to measure how effectively representations learned from the source language transfer to the target language under different tokenization schemes.

\subsection{Evaluation on Downstream POS Task}
We evaluate tokenizer performance through POS tagging, a controlled extrinsic task that reflects how well subword segmentation supports morphosyntactic learning in agglutinative Uralic languages. To ensure robustness across architectures, we employ two complementary sequence labeling models: a BiLSTM–CRF \citep{lample-etal-2016-neural, ma-hovy-2016-end} and Flair \citep{akbik2018coling, akbik2019flair}. The BiLSTM–CRF relies solely on local contextual representations learned from subword embeddings, providing a clean testbed for segmentation sensitivity, while Flair uses character-level contextual string embeddings that capture long-range dependencies and subword-internal regularities, which is particularly advantageous for morphologically rich languages. 

We report Accuracy and Macro-F1 as evaluation metrics. Accuracy reflects overall tagging correctness, whereas Macro-F1 assigns equal weight to all POS categories, preventing frequent tags from dominating the evaluation. Together, these metrics allow us to assess how tokenizer design influences both general and category-sensitive POS performance.

\section{Results}
\label{sec:ud-eval}

\begin{table*}[t]
\centering
\small
\renewcommand{\arraystretch}{1.15}
\resizebox{0.8\linewidth}{!}{
\begin{tabular}{l l c l | c c c c}
\hline
\textbf{Source} & \textbf{Target} & \textbf{Sim.} & \textbf{Tokenizer}
& \multicolumn{2}{c}{\textbf{BiLSTM--CRF}} & \multicolumn{2}{c}{\textbf{Flair}} \\
\cline{5-8}
& & & & Acc & Mac F1 & Acc & Mac F1 \\
\hline

\multirow{3}{*}{\latin{} \texttt{est}} & \multirow{3}{*}{\texttt{hun}} & \multirow{3}{*}{✗}
& BPE     & 0.8096 & 0.7013 & 0.9509 & \textcolor{blue}{0.7930} \\
& & & Unigram & 0.7840 & 0.6663 & 0.9408 & 0.7651 \\
& & &  OBPE
& \textcolor{blue}{0.8496} & \textcolor{blue}{0.7398} & \textcolor{blue}{0.9614} & 0.7902 \\
\hline

\multirow{3}{*}{\latin{} \texttt{est}} & \multirow{3}{*}{\texttt{sme}} & \multirow{3}{*}{\textcolor{green!50!black}{\checkmark}}
& BPE     & 0.7749 & 0.7573 & 0.9075 & 0.8050 \\
& & & Unigram & 0.7830 & 0.7573 & 0.9078 & 0.7885 \\
& & &  OBPE
& \textcolor{blue}{0.8152} & \textcolor{blue}{0.7850} & \textcolor{blue}{0.9373} & \textcolor{blue}{0.8390} \\
\hline

\multirow{3}{*}{\latin{} \texttt{fin}} & \multirow{3}{*}{\texttt{hun}} & \multirow{3}{*}{✗}
& BPE     & 0.8096 & 0.7013 & 0.9509 & \textcolor{blue}{0.7930} \\
& & & Unigram & 0.7840 & 0.6663 & 0.9408 & 0.7651 \\
& & &  OBPE
& \textcolor{blue}{0.8514}
& \textcolor{blue}{0.7412}
& \textcolor{blue}{0.9581}
& 0.7907 \\
\hline

\multirow{3}{*}{\latin{} \texttt{fin}} & \multirow{3}{*}{\texttt{sme}} & \multirow{3}{*}{\textcolor{green!50!black}{\checkmark}}
& BPE     & 0.7749 & 0.7573 & 0.9075 & 0.8050 \\
& & & Unigram & 0.7830 & 0.7573 & 0.9078 & 0.7885 \\
& & &  OBPE
& \textcolor{blue}{0.8036}
& \textcolor{blue}{0.7914}
& \textcolor{blue}{0.9264}
& \textcolor{blue}{0.8164}  \\
\hline

\multirow{3}{*}{\cyril{} \texttt{rus}} & \multirow{3}{*}{\texttt{kpv}} & \multirow{3}{*}{✗}
& BPE     & 0.6744 & 0.4742 & 0.3941 & 0.4409 \\
& & & Unigram
& \textcolor{blue}{0.7401}
& \textcolor{blue}{0.5209}
& \textcolor{blue}{0.9101}
& \textcolor{blue}{0.6367} \\
& & &  OBPE
& 0.7207
& 0.5022
& 0.8930
& 0.5693 \\
\hline
\end{tabular}
}
\caption{
POS tagging performance across source--target pairs. Target pairs with North Sámi (\textcolor{green!50!black}{\checkmark}) exhibit high linguistic similarity.
}
\label{tab:pos_model_comparison}
\end{table*}

\paragraph{The general performance.}
Table~\ref{tab:pos_model_comparison} reports tokenizer performance on POS tagging. OBPE generally surpasses both BPE and Unigram baselines in terms of accuracy and F1 scores in most settings. The main exception is the Cyrillic group, where Unigram consistently performs better. In addition, Flair achieves stronger results with BPE on Hungarian. These observations will be examined in more detail later.

\paragraph{The long-tail issue.}
A marked discrepancy between Accuracy and Macro-$F1$ in the Hungarian results highlights a structural limitation of frequency-based segmentation. Although standard BPE attains high Accuracy (0.8096), its Macro-$F1$ (0.7013) lags substantially behind OBPE (0.7412). This disparity illustrates the \textit{long-tail} challenge in agglutinative morphology: a small set of central categories dominates the corpus, while numerous complex inflected forms occur only rarely. Because standard BPE seeks to minimize the corpus description length, it preferentially segments the frequent stems of the dominant classes, thus achieving strong frequency-weighted accuracy \citep{gutierrez-vasques-etal-2023-languages}.

Since Macro-$F1$ weights all classes equally, independent of how frequent they are, it penalizes models that fail to generalize to rare morphological patterns. By contrast, OBPE uses cross-lingual anchors to better maintain these infrequent affix boundaries, resulting in a higher Macro-$F1$ that reflects strong performance across the entire range of morphological detail.

\paragraph{Cyrillic performance gap.}
We attribute the performance gap in the Cyrillic group primarily to the substantial linguistic distance between the source (Russian) and target (Komi-Zyrian) languages. While centuries of contact have introduced Russian loanwords into Komi-Zyrian, the core grammatical inventory remains distinct \citep{69dacb55baa44bb2a5a0ab88631ea71a}. There is structural discordance between the fusional morphology of Russian and the agglutinative typology of Komi-Zyrian. Russian suffixes often encode case, number, and gender into a single fused unit; for instance, the ending \textit{-am} in \textit{stol-\textbf{am}} \footnote{Cyrillic examples are transliterated using the transliteration toolset provided by the COPIUS initiative: \url{https://www.copius.eu/ortho.php}.}(tables) simultaneously marks dative case and plural number. In contrast, Komi-Zyrian concatenates distinct suffix chains for each category in the form of Root-Num-Case, as seen in \textit{pezan-\textbf{jas}-\textbf{li}}, where the plural (\textit{-jas}) and dative (\textit{-li}) markers remain distinct segments. Furthermore, Russian relies heavily on free-standing prepositions to express spatial relations, whereas Komi-Zyrian encodes these relations exclusively through bound case suffixes like most Uralic languages. This asymmetry creates a bottleneck for OBPE: the overlap objective cannot find the shared morphological anchors despite the shared script. Thus, the shared Cyrillic script acts as a "false friend". 

\begin{table*}[t]
\centering
\footnotesize
\setlength{\tabcolsep}{3pt}
\renewcommand{\arraystretch}{1.06}

\resizebox{\textwidth}{!}{%
\begin{tabular}{l|
cccc|cccc|
ccc|ccc|
cccc|cccc}
\hline

& \multicolumn{8}{c|}{\textbf{\texttt{sme}}}
& \multicolumn{6}{c|}{\textbf{\texttt{kpv}}}
& \multicolumn{8}{c}{\textbf{\texttt{hun}}} \\

\textbf{POS}
& \multicolumn{4}{c}{\textbf{BiLSTM}} & \multicolumn{4}{c|}{\textbf{Flair}}
& \multicolumn{3}{c}{\textbf{BiLSTM}} & \multicolumn{3}{c|}{\textbf{Flair}}
& \multicolumn{4}{c}{\textbf{BiLSTM}} & \multicolumn{4}{c}{\textbf{Flair}} \\

& BPE & Uni & OFin & OEst & BPE & Uni & OFin & OEst
& BPE & Uni & OBPE & BPE & Uni & OBPE
& BPE & Uni & OFin & OEst & BPE & Uni & OFin & OEst \\

\hline
ADJ
& 0.517&0.583&\textbf{\textcolor{red}{0.644}}&0.637&0.638&0.594&0.718&0.667
& 0.412&0.409&0.350&0.449&0.427&0.464
& 0.627&0.789&\textbf{\textcolor{red}{0.732}}&0.716&0.835&0.789&0.830&0.834 \\

ADP
& 0.750&0.717&0.764&0.767&0.829&0.728&0.749&0.772
& 0.425&0.595&0.587&0.600&0.675&0.590
& 0.856&0.907&0.888&0.860&0.920&0.907&0.897&0.905 \\

ADV
& 0.753&0.753&0.732&0.748&0.822&0.782&0.789&0.784
& 0.597&0.654&0.639&0.696&0.731&\textbf{\textcolor{red}{0.722}}
& 0.774&0.754&0.822&\textbf{\textcolor{red}{0.806}}&0.866&0.833&0.868&0.850 \\

AUX
& 0.785&0.789&0.791&0.810&0.807&0.811&0.811&0.825
& 0.737&0.768&0.721&0.775&0.777&0.807
& 0.750&0.792&0.816&0.779&0.812&0.750&0.765&0.808 \\

CCONJ
& 0.987&0.968&0.990&0.984&0.990&0.958&0.967&0.977
& 0.798&0.732&0.834&0.798&0.850&0.859
& 0.939&0.919&0.949&0.947&0.944&0.939&0.947&0.927 \\

INTJ
& 0.857&0.857&0.857&0.667&0.400&0.667&0.667&0.857
& 0.000&0.000&0.000&0.000&0.000&0.000
&  -- & -- & -- & -- & -- & -- & -- & -- \\

NOUN
& 0.720&0.739&\textbf{\textcolor{red}{0.768}}&0.784&0.809&0.769&0.810&0.831
& 0.587&0.679&0.632&0.298&0.766&\textbf{\textcolor{red}{0.734}}
& 0.844&0.736&0.812&0.811&0.887&0.844&0.876&0.882 \\

NUM
& 0.630&0.563&0.594&0.646&0.706&0.671&0.667&0.740
& 0.455&0.556&0.556&0.400&0.476&0.133
& 0.895&0.832&0.736&0.809&0.895&0.833&0.810&0.860 \\

PART
& 0.699&0.750&0.805&0.778&0.838&0.773&0.813&0.803
&  -- & -- & -- & -- & -- & --
& 0.857&0.490&0.867&0.903&0.857&0.903&0.968&0.933 \\

PRON
& 0.922&0.894&0.912&0.931&0.948&0.933&0.947&0.943
& 0.758&0.802&0.757&0.629&0.893&0.809
& 0.681&0.673&0.713&0.682&0.781&0.732&0.806&0.776 \\

PROPN
& 0.589&0.504&0.665&0.680&0.677&0.612&0.690&0.741
& 0.000&0.000&0.000&0.000&0.000&0.000
& 0.781&0.670&0.816&0.833&0.880&0.811&0.895&0.904 \\

PUNCT
& 0.999&0.997&0.995&0.998&0.999&0.994&0.993&0.999
& 0.992&0.985&0.994&0.997&0.997&0.985
& 0.998&0.992&0.997&0.997&0.998&0.997&0.991&0.996 \\

SCONJ
& 0.742&0.798&0.859&0.840&0.857&0.808&0.861&0.868
& 0.829&0.739&0.800&0.571&0.857&0.737
& 0.953&0.911&0.949&0.928&0.953&0.902&0.941&0.922 \\

VERB
& 0.653&0.693&0.704&0.721&0.761&0.731&0.767&0.780
& 0.523&0.695&\textbf{\textcolor{red}{0.665}}&0.375&0.772&\textbf{\textcolor{red}{0.780}}
& 0.891&0.708&0.799&0.804&0.891&0.805&0.888&0.873 \\

\hline
\textbf{Macro Avg}
& 0.757&0.757&0.791&0.785&0.805&0.789&0.816&0.839
& 0.508&0.558&0.538&0.441&0.6367&0.5693
& 0.701&0.666&0.741&0.740&0.793&0.765&0.790&0.790 \\
\hline
\end{tabular}
}
\caption{POS-wise F1 across three cross-lingual settings (\texttt{sme, kpv, hun}). OFin/OEst denote OBPE trained on Finnish/Estonian.}
\label{tab:poswise}
\end{table*}

\paragraph{Category level effects} 

We report detailed per-tag metric scores in Table~\ref{tab:poswise}. The results indicate that tokenizer-related discrepancies occur mainly in open-class, morphologically productive tag categories. 

In North Sámi, switching from BPE to OBPE leads to the largest improvements in morphologically complex open-class categories. For ADJ, BPE attains an F1 of just 0.5173, while OBPE (sme-fi) increases this to 0.6440. NOUN tagging similarly benefits, rising from 0.7200 (BPE) to 0.7681 (OBPE). These gains reflect the differing vocabulary construction goals of the two methods.

Functional categories like PUNCT and CCONJ remain highly stable across tokenizers ($F1>0.98$ for North Sámi). \textbf{This indicates that tokenization choice mainly affects morphologically rich categories, not fixed-vocabulary function classes.} Since both BPE and Unigram are designed to keep high-frequency items intact in the vocabulary \citep{sennrich-etal-2016-neural, kudo-2018-subword}, these closed-class tokens are almost always encoded as single units in every model, leading to consistently similar performance.

The limitations of BPE in extremely low-resource conditions are clearly illustrated by the Komi-Zyrian results in Table~\ref{tab:poswise}. The BPE model performs disastrously on VERB ($F1=0.375$), while the OBPE model captures substantially more verbal structure ($F1=0.780$). This aligns with \citet{bostrom-durrett-2020-byte} showing that BPE's deterministic merge operations are ill-suited to settings where the available data are too sparse to derive reliable frequency statistics for complex morphologies.

\section{Further Study}
Although dataset size sets a basic upper bound on model performance, our analysis shows that Tag Diversity plays an important compensatory role. To operationalize this notion, we computed POS Entropy ($H$) from the tag distribution in the training data (see Table \ref{tab:bilstm_results_by_trainsize} for summary statistics). Specifically, for a tag set $T$ and tag probabilities $P(t)$:
\begin{equation}
H(T) = - \sum_{t \in T} P(t) \log_2 P(t)
\end{equation}
Higher $H$ values correspond to a more even spread of grammatical categories, thereby providing wider morphological coverage.

Hungarian illustrates this relationship clearly. Despite its dataset being roughly half the size of the North Sámi corpus (910 vs. 1,873 sentences), it exhibits higher syntactic entropy ($H \approx 2.275$ vs. $2.195$). As a result, the performance degradation is relatively modest given the data reduction: OBPE attains a Macro-$F1$ of 0.741 on Hungarian, remaining close to the North Sámi score of 0.815 even with 50\% fewer training instances. In contrast, Komi (kph), which is affected by both severe data sparsity (397 sentences) and the lowest entropy ($H \approx 2.018$), shows a pronounced performance loss (Unigram $F1$: 0.521). These findings suggest that languages with richer and more evenly distributed tag inventories enable tokenizers to generalize better under limited supervision, while low-entropy distributions intensify the impact of data scarcity.

\begin{table}[!t]
    \centering
    \setlength{\tabcolsep}{10pt} 
    \resizebox{\linewidth}{!}{
    \begin{tabular}{lccc}
        \toprule
        \textbf{Tokenizer} & \textbf{sme} & \textbf{hun} & \textbf{kph} \\
        \midrule
        \textit{Train size / POS Ent.} & \textit{1873 / 2.195} & \textit{910 / 2.275} & \textit{397 / 2.018} \\
        \midrule
        BPE        & \textbf{0.757}& 0.701 & 0.674 \\
        Unigram    & \textbf{0.783} & 0.666 & 0.521 \\
        OBPE (fi)  & \textbf{0.804} & 0.741 & -- \\
        OBPE (et)  & \textbf{0.815} & 0.7398 & -- \\
        OBPE (rus) & --     & --     & 0.5022 \\
        \bottomrule
    \end{tabular}
    }
    \caption{BiLSTM Macro-$F1$ across languages. The top section details training size and POS entropy for each language.}
    \label{tab:bilstm_results_by_trainsize}
\end{table}

\begin{table*}[h]
    \centering
    \small
    \setlength{\tabcolsep}{3pt}
    \resizebox{0.7\textwidth}{!}{
    \begin{tabular}{l l l}
    \toprule
    \textbf{Phenomenon} & \textbf{Segmentation} & \textbf{Gloss} \\
    \midrule

    \multirow{2}{*}{Agglutination} & \textit{szervezet-e-i-re} & ``to its organizations'' \\
    & \footnotesize{(\textsc{org}+\textsc{poss}+\textsc{pl}+\textsc{case})} & \\
    \midrule

    \multirow{2}{*}{Vowel Harmony} & \textit{globalizáció} + \textit{-ra} & ``to globalization'' \\
     & \textit{rendőrség} + \textit{-re} & ``to the police'' \\
    \midrule
    \multirow{2}{*}{Consonant Gradation} & \textit{veahkki} (lemma)& ``help'' \\
    & \textit{veahki} & \\
    \bottomrule
    \end{tabular}
    }
    \caption{Typological challenges extracted from the train corpora: agglutination, vowel harmony, and consonant gradation.}
    \label{tab:morph_examples}
\end{table*}
\section{Qualitative Analysis}
We present three categories of morphologically challenging examples from Uralic languages and show how OBPE addresses them effectively compared to other approaches.
\begin{itemize}
\item \textbf{Agglutination: } a process where complex words are formed by concatenating distinct suffixes to a single root stem. As illustrated in Table~\ref{tab:morph_examples}, a single Uralic token often conveys information that would require an entire phrase in English. For instance, the Hungarian token \textit{szervezeteire} exhibits a massive accumulation of suffixes: it combines the root \textit{szervezet} (organization) with the possessive marker \textit{-e}, the plural marker \textit{-i}, and the case marker \textit{-re} This morphological density results in an explosion of word forms. This poses a severe challenge for frequency-based tokenizers like BPE: the specific combination may appear only once in the corpus, preventing the model from learning a stable representation for the frequent constituent parts.
\item \textbf{Vowel Harmony:} Suffixes must match the harmonic class (front/back) of the root. This forces the tokenizer to learn multiple allomorphs (e.g., \textit{-ban} vs. \textit{-ben}). Crucially, this affects tokenization by decreasing the statistical signal of grammatical markers.
\item \textbf{Consonant Gradation:} In contrast to harmony, North Sami exhibits consonant gradation, where the stem's internal consonants change length or quality based on the grammatical context. For instance, the lemma \textit{veahkki} ("help") appears as \textit{veahki} in the accusative case (see Table~\ref{tab:morph_examples}). This stem alternation disrupts the static subword patterns that BPE relies on for consistent root recognition.
\end{itemize}

As in Table~\ref{tab:morph_examples}, the Hungarian token \textit{szervezeteire} encapsulates a root plus three distinct morphemes. This structure creates a explosion of word forms, resulting in a "long tail" where rare suffix combinations evade the frequency thresholds of standard tokenizers like BPE, leading to over-segmentation.

\section{Conclusion}
This study presented a controlled evaluation of subword tokenization strategies across six Uralic languages, clarifying how morphology and resource imbalance shape tokenization behavior. Our findings reveal that tokenization is not a neutral preprocessing step but \textbf{a decisive factor in low-resource and morphologically rich contexts}.

Three major insights emerge. First, genealogical proximity amplifies cross-lingual transfer: Overlap BPE (OBPE) achieves substantial gains when low-resource languages share a structural base with their source languages (e.g., Finnish–North Sámi), but not when such alignment is absent (e.g., Russian–Komi-Zyrian). Second, in isolated low-resource settings, the Unigram model provides more faithful morphological segmentation than BPE, owing to its probabilistic pruning and subword regularization. Third, these improvements are concentrated in open-class categories—nouns and verbs—where morphological complexity most challenges standard segmentation.

Taken together, our findings indicate that standard BPE is often ill-suited for Uralic and other highly agglutinative language families, a limitation that is likely to become more pronounced in real-world multilingual applications. As a practical guideline, we recommend OBPE in genealogically grounded multilingual settings, and Unigram in isolated low-resource scenarios. This provides a simple yet effective strategy for morphology-sensitive tokenization in multilingual NLP.

\section*{Limitations}
\label{sec:limitations}
Our research concentrates on a particular subset of the Uralic language family, which constrains the direct applicability of our results to other, typologically different language groups. In addition, the very limited amount of labeled data—especially for Komi-Zyrian (397 sentences)—reduces the statistical reliability of our experiments and makes extensive cross-validation infeasible. 

To address the severe data sparsity in Komi-Zyrian, we fixed the tokenizer vocabulary size to 5{,}000 subword units. We recognize that this is substantially smaller than the typical vocabulary sizes (30k or more) used in contemporary Large Language Models (LLMs). Accordingly, it remains unresolved whether the morphological benefits of OBPE would hold when scaled up to the much larger settings required for open-domain text generation.

Our evaluation is limited to POS tagging, a relatively "shallow" syntactic task that primarily relies on local contextual signals. Future research must determine whether our conclusions also extend to more "deep" semantic tasks such as NLI or Machine Translation, where subword segmentation has a stronger impact on meaning-bearing elements. 

Lastly, we highlight a possible domain bias in our cross-lingual transfer configuration. The high-resource source models are trained on newswire text, whereas the low-resource Komi-Zyrian corpus contains only fictional prose. This domain mismatch acts as a confound, since the models must simultaneously adapt to a new language and a different textual genre.

\section*{Discussion}
\label{sec:discussion}
While we attribute the primary performance gap to typological distance, we acknowledge that the distinct scripts introduce a confounding variable that is difficult to isolate in the current setup. A rigorous disentanglement of orthographic opacity from morphological incompatibility would require projecting the Cyrillic data into a shared phonemic space (e.g., via UPA or IPA transliteration). Such a control would isolate the morphological alignment process from script-specific artifacts, thereby clarifying whether the limitations of OBPE in this setting are intrinsic to the linguistic mismatch or exacerbated by orthographic divergence.

Besides, we must contextualize our results within the landscape of practical NLP. While OBPE outperforms baselines, the absolute scores for low-resource targets remains far below "production-ready" standards. For comparison, the top-performing system for English in the CoNLL 2018 Shared Task achieved a UPOS $F1$ of 0.959 \citep{zeman-etal-2018-conll}. This indicates that achieving reliable utility for downstream tasks in extreme low-resource settings will likely require complementary transfer techniques, such as adapters or syntactic projection, to bridge the gap between statistical significance and industrial usability.

\section*{Acknowledgment}
This research was supported by G-LAMP Program of the National Research Foundation of Korea (NRF) grant funded by the Ministry of Education (No. RS-2025-25441317).

\bibliography{anthology.min,custom}

\end{document}